\documentclass[10pt]{article}
\usepackage[letterpaper]{geometry}
\usepackage{hicss}
\usepackage{times}
\usepackage[none]{hyphenat}
\usepackage{url}
\usepackage{latexsym}
\usepackage{indentfirst}
\usepackage{graphicx}
\graphicspath{{images/}}

\usepackage[hidelinks]{hyperref}

\usepackage{tabularx}  % To have the table fill out the page
\usepackage{cleveref}

\usepackage{caption}
\usepackage{subcaption}

\usepackage{enumitem}

\usepackage{amsthm}

\newtheoremstyle{requirement_style}          % name
{5pt}                                       % space above
{5pt}                                        % space below
{\itshape}                                   % body font
{14pt}                                       % indent amount
{\bfseries}                                  % theorem head font
{:}                                          % punctuation after theorem head
{0.5em}                                      % space after theorem head
{\thmname{#1}\thmnumber{#2}\thmnote{ (#3)}}  % theorem head spec

\theoremstyle{requirement_style}
\newtheorem{requirement}{DR}
\newtheorem{principle}{DP}

\title{Deep Learning Strategies for Industrial Surface Defect Detection Systems}

\author{Dominik Martin \\
 Karlsruhe Institute\\ of Technology (KIT) \\
 {\small{\underline{dominik.martin@kit.edu}}} \\\And
 Simon Heinzel \\
 Karlsruhe Institute\\ of Technology (KIT) \\
 {\small{\underline{simon.heinzel@alumni.kit.edu} }}\\\And
 Johannes Kunze von Bischhoffshausen \\
 Trelleborg Sealing Solutions\\ Germany GmbH \\
 {\small{\underline{johannes.kunze@trelleborg.com} }}\\\And 
 Niklas Kühl \\
 Karlsruhe Institute\\ of Technology (KIT) \\
 {\small{\underline{niklas.kühl@kit.edu}}} \\}

\date{}

\begin{document}
\maketitle
\begin{abstract}
Deep learning methods have proven to outperform traditional computer vision methods in various areas of image processing.
However, the application of deep learning in industrial surface defect detection systems is challenging due to the insufficient amount of training data, the expensive data generation process, the small size, and the rare occurrence of surface defects.
From literature and a polymer products manufacturing use case, we identify design requirements which reflect the aforementioned challenges. 
Addressing these, we conceptualize design principles and features informed by deep learning research.
Finally, we instantiate and evaluate the gained design knowledge in the form of actionable guidelines and strategies based on an industrial surface defect detection use case.
This article, therefore, contributes to academia as well as practice by (1) systematically identifying challenges for the industrial application of deep learning-based surface defect detection, (2) strategies to overcome these, and (3) an experimental case study assessing the strategies' applicability and usefulness.
\end{abstract}

\section{Introduction}
\label{sec:introduction}

% Motivation/Relevance/Significance
Quality control is an essential process in the manufacturing industry \cite{Ferguson2018}. As part of quality management, it ensures the quality of manufactured products. In this process, the visual inspection of finished products plays an important role \cite{Tabernik2019}. Typically, this task is carried out manually, and workers are trained to identify complex surface defects \cite{Beyerer2015}. However, manual visual inspection is monotonous, laborious, fatiguing, subjective, lacking in good reproducibility, too slow in many cases, and costly. As a result, automated visual inspection systems have spread in the industry since the 1970s. The main benefits of such systems include impartial and reproducible inspection results, complete and detailed documentation, faster inspection rates, and lower costs \cite{Beyerer2015, Tabernik2019}.

In the past, these systems relied on traditional computer vision methods, which addressed at least some of the issues of manual visual inspection \cite{Xie2008}. 
However, with the Industry 4.0 paradigm, which aims to increase automation of traditional manufacturing processes through digitization, the trend is moving towards the generalization of the production line, where rapid adaptation to a new product is required \cite{Oztemel2018}. 
Traditional computer vision methods are unable to provide such flexibility. They rely on a two-step process of extracting handcrafted features and training an appropriate classifier. The critical step in this process lies in the extraction of robust handcrafted feature representations for the specific problem at hand \cite{Weimer2016}. This step leads to lengthy development cycles \cite{Tabernik2019} and requires a high level of human expertise \cite{Ren2017}.
A solution that allows for improved flexibility and reduced engineering efforts can be found in deep learning methods. Deep learning methods learn the relevant features directly from the raw data, eliminating the need for handcrafted feature representations. In recent years, these methods have reached and even exceeded human-level performance on image-related tasks such as image classification \cite{Aggarwal2018}.

% Problem
However, deep learning methods are still rarely applied in automated visual inspection systems due to several reasons \cite{ZschechWalkHeinrich2021_1000131822}. The available datasets are usually too small to train deep neural networks \cite{Soukup2014, Weimer2016, Park2016, Ren2017, Ferguson2018, Tao2018, Di2019} and the generation of such datasets is expensive due to the intensive manual work required for labeling the data \cite{Ren2017, Tao2018}. Additionally, surface defects can be extremely small, making their detection even more challenging \cite{Park2016, Wang2016}. The black-box nature of deep neural networks also makes it difficult for human domain experts to understand what the network considers a defect \cite{Racki2018}. 

Against this background, we contribute to the information systems (IS) literature by investigating suitable strategies that enable the successful application of deep learning methods in industrial surface defect detection systems (SDDS). More specifically, we aim to answer the following research questions:

% Research questions
\begin{enumerate}[label=\bfseries RQ\arabic*:, leftmargin=*, labelindent=0pt]
	\item Which challenges exist for deep learning methods in industrial SDDS, and which design requirements can be derived from these challenges?
	\item Which deep learning strategies in the form of design principles and design features address these design requirements and are suitable for industrial SDDS?
	\item Which strategies achieve the best performance in industrial SDDS?
\end{enumerate}

\section{Research Design}
\label{sec:research_design}

To address the research questions raised, we follow the Design Science Research (DSR) paradigm \cite{Gregor2007, March1995}. 
Overall, we base our research on the three cycle view proposed by Hevner \cite{Hevner2007}, which ensures practical applicability on the one hand and rigorous construction and evaluation of innovative artifacts on the other.
Thus, we aim to create artifacts that solve the problems of a specific application domain (relevance cycle) while drawing on applicable knowledge from theory (rigor cycle). 
In this particular research, we contribute to the application domain of surface defect detection in the manufacturing industry and base our artifact construction on literature from the field of deep learning.

\begin{figure*}[h]
	\centering
	\includegraphics[width=1\textwidth]{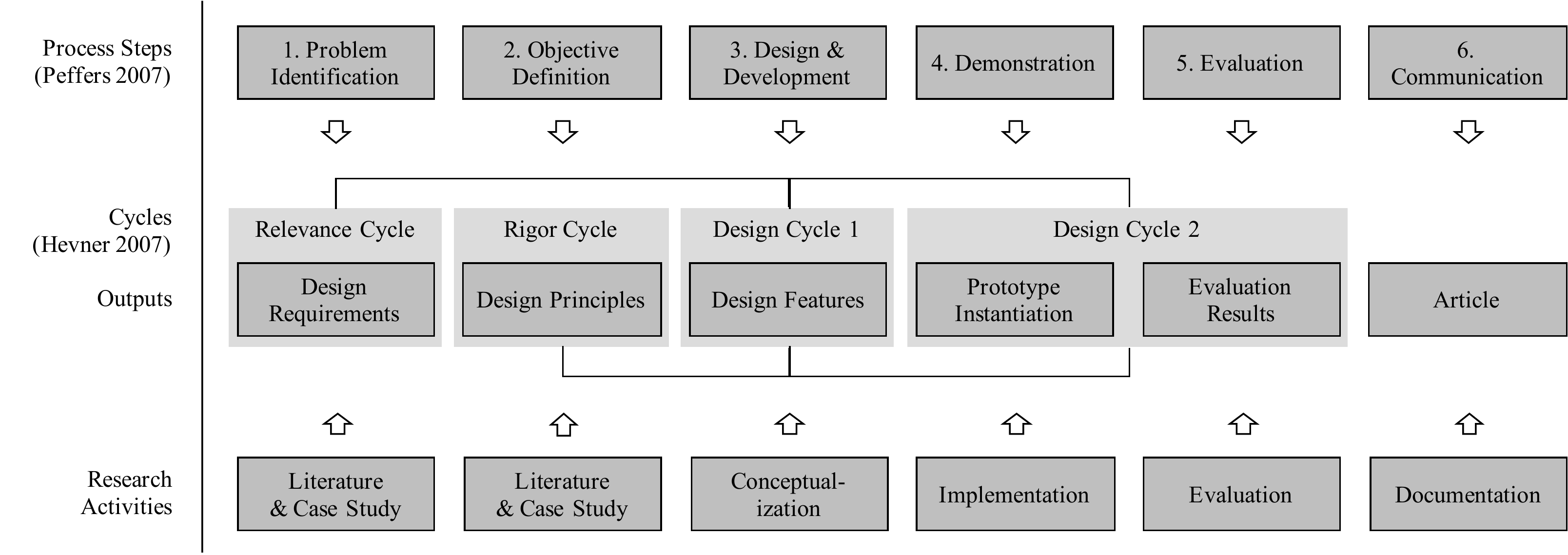}
	\caption{Overall research design based on Peffers et al. \cite{Peffers2007} and Hevner \cite{Hevner2007}}
	\label{fig:research_design}
\end{figure*}

Our specific approach is based on the DSR process model presented by Peffers et al. \cite{Peffers2007} and consists of six subsequent steps. \Cref{fig:research_design} illustrates these steps, the resulting outputs, and the corresponding research activities.
% Step 1
First, we define the research problem by identifying domain-specific challenges for deep learning methods in relevant literature as well as through exploratory focus groups in a case company \cite{yin2011applications}. In several focus group sessions conducted, seven experts from different areas such as operations, quality control and data science were involved. From the challenges identified, we derive \emph{design requirements} (DR), which represent generic requirements that should be met by any artifact aiming to solve these problems \cite{Chanson2019}. This step corresponds to the relevance cycle and addresses \emph{RQ1}.
% Step 2
Second, we define the objectives of a solution by inferring \emph{design principles} (DP) from the design requirements. Design principles are generic capabilities of an artifact through which the design requirements are addressed \cite{Chanson2019}. We base the design principles on relevant literature from the field of deep learning; hence this step corresponds to the rigor cycle.
% Step 3
In the third step, we derive \emph{design features} (DF) that address the design principles and conceptualize a framework of interrelated design requirements, design principles, and design features. Design features are specific capabilities of an artifact that fulfill and implement the design principles \cite{Chanson2019}. A design principle that is instantiated by a design feature can be understood as an explanation (design principle) of why a specified piece (design feature) leads to a predefined goal (design requirement). This step corresponds to the first design cycle and, together with the previous step, answers \emph{RQ2}.
% Step 4
Next, we validate the artifact proposed in the first design cycle and demonstrate its feasibility, applicability and usefulness \cite{pries2008strategies} by instantiating it in the context of an exemplary surface defect detection use case. We conduct eight experiments leveraging strategies from the framework.
% Step 5
In step five, we evaluate the deep learning models and draw conclusions about the different deep learning strategies. Steps four and five, thus, address \emph{RQ3} and represent the second design cycle.
% Step 6
Finally, we contribute to the body of knowledge by communicating the identified challenges, the created artifacts, and the evaluation results in the article at hand.

% Figure: Research design

In summary, the first artifact is a \emph{framework} of interrelated design requirements, principles, and features which captures suitable deep learning strategies for enabling industrial surface defect detection systems. The second artifact is an \emph{instantiation} of the framework on an industrial use case in the field of visual inspection of engineered molded parts.
In a series of experiments, we build different deep learning models leveraging strategies from the framework illustrating their feasibility, applicability and usefulness. 
Thus, this article aims to contribute design knowledge in the form of \emph{operational principles/architectures} and a \emph{situated implementation of an artifact} \cite{Gregor2013}. Hence, it makes a \emph{level 2} (design cycle 1) and \emph{level 1} (design cycle 2) contribution according to Gregor and Hevner \cite{Gregor2013}. The DSR knowledge contribution type represents an \emph{exaptation}, since this article aims to extend known solutions to new problems \cite{Gregor2013}.

\section{Relevance Cycle: Design Requirements for Surface Defect Detection Systems}
\label{sec:relevance}

The relevance cycle aims to place the research in a contextual environment and provide requirements and acceptance criteria for the design science activities. Thus, we provide an overview of previous research on surface defect detection in the manufacturing industry to identify domain-specific challenges for deep learning methods. By leveraging insights from related literature as well as an exploratory case study with experts from industry, identified challenges are condensed into design requirements.

\subsection{Surface Defect Detection}
\label{sec:surface_defect_detection}

The detection of surface defects using computer vision techniques has been widely studied in the literature. Surface defects are considered local anomalies in homogeneous textures like scratches, cracks, holes, etc. This includes a wide range of surface textures, including textile \cite{Murino2004}, wood \cite{Silven2003}, metal \cite{Pernkopf2004} and ceramic tiles \cite{Xie2007}. The methods commonly leveraged can be divided into \emph{traditional computer vision methods} and \emph{deep learning methods}. Traditional computer vision methods are based on a two-step process of extracting handcrafted features and training an appropriate classifier such as an SVM or decision tree. The critical step in this process lies in the extraction of robust handcrafted feature representations for the specific problem at hand.

Xie \cite{Xie2008} categorizes the methods used to extract these features into four different approaches:
\emph{Structural approaches} focus on texture elements and their spatial arrangement. They extract texture primitives such as simple line segments, individual pixels, or regions with uniform gray-levels and generate a dynamic texture model by applying some spatial placement rules. Structural methods are usually applied to repetitive patterns such as textile \cite{Chen1988}, fabrics \cite{MallikGoswami2000}, and leather \cite{Wen1999}. Popular structural approaches include primitive measurement \cite{Kittler1994}, edge features \cite{Wen1999}, skeleton representation \cite{Chen1988}, and morphological operations \cite{MallikGoswami2000}.
\emph{Statistical approaches} analyze the spatial distribution of pixel values. They work well on stochastic textures, such as ceramic tiles, castings, and woods. In this category, researchers use numerous statistics, such as histogram properties \cite{Kim1994}, co-occurrence matrices \cite{Conners1983}, local binary patterns \cite{Niskanen2001}, autocorrelation \cite{Zhu2015} and others.
\emph{Filter-based approaches} apply filters to detect features, such as edges, textures, and regions. They can be further divided into spatial domain filtering \cite{Ade1984}, frequency domain filtering \cite{Ravandi1995}, and spatial-frequency domain filtering \cite{Hu2016}.
\emph{Model-based approaches} construct representations of images by modeling multiple properties of the defects. In this category, researchers use fractal models \cite{Conci1998}, autoregressive models \cite{Serafim1991}, and random field models \cite{Cohen1991}.

Shortly after the introduction of AlexNet \cite{Krizhevsky2012}, deep learning methods began being applied more often to surface defect detection problems. The motivation arises from the difficulty that even domain experts struggle to design the right set of features to detect certain defects.
Masci et al. \cite{Masci2012} show that deep learning methods can significantly outperform traditional computer vision methods. They use a CNN consisting of five layers for the classification of steel defects and achieved excellent results.
The work from Soukup and Huber-Mörk \cite{Soukup2014} shows that regularization methods like unsupervised layer-wise pre-training and data augmentation yield further performance improvements. %They demonstrate this on the classification of rail surface defects using a CNN consisting of four layers.
Weimer et al. \cite{Weimer2016} evaluate several deep learning architectures with varying depths and widths of layers on a synthetic texture dataset. %They show that a larger depth and width of the network has a positive impact on classification results.
The work from Ren et al. \cite{Ren2017} shows that using a pre-trained network improves the performance of deep learning methods. They also extend the problem of surface defect detection from image classification to image segmentation.
%\cite{Racki2018} proposed a network for explicitly performing segmentation of defects. They implemented a fully convolutional network with 10 layers, using both ReLU and batch normalization to perform the segmentation of the defects. Furthermore, they proposed an additional decision network on top of the features from the segmentation network to perform a per-image classification of a defects presence. This allowed them to improve the classification accuracy on the dataset of synthetic surface defects.

\subsection{Design Requirements}
\label{sec:design_requirements}

However, especially the application of deep learning approaches opens up a number of previously inadequately explored challenges.
% DR1
One challenge, pointed out by several authors, is the particularly small size of the defects \cite{Park2016, Wang2016}. Also in our selected industrial use case, the defects are so small that they are difficult to see with the naked eye. This makes it more difficult to detect the defects and capture them in a way that the defects are also visible in the images. Consequently, we derive the following design requirement:
\begin{requirement}%[Detection of very small defects]
	Industrial surface defect detection systems should be able to detect very small defects.
\end{requirement}

% DR2
A second challenge lies in the rare occurrence of defects \cite{Ren2017, Shang2018, Di2019}. Datasets from manufacturing processes are often highly imbalanced due to the deliberately low probability of defect occurrences. Deep learning methods in general are designed to minimize the overall loss, which can result in paying more attention to the majority class and not properly learning the appearance of the minority class. Consequently, this issue has to be addressed appropriately:
\begin{requirement}%[Detection of rarely occurring defects]
	Industrial surface defect detection systems should  be able to detect rarely occurring defects.
\end{requirement}

% DR3
A third challenge concerns the difficulty in understanding deep neural networks \cite{Racki2018}. Deep neural networks are black-box networks, making them difficult to understand or interpret \cite{Dargan2019}. The quality inspectors in our use case also emphasize the importance of trusting deep learning methods because their model decisions as such are untraceable; thus:
\begin{requirement}%[Explainability of model decision]
	The decisions of industrial surface defect detection systems should be explainable.
\end{requirement}

% DR4
A fourth challenge is the insufficient amount of training data. Several authors point out that the size of datasets is usually too small to train deep neural networks and that the training is prone to overfitting \cite{Soukup2014, Weimer2016, Park2016, Ren2017, Ferguson2018, Tao2018, Di2019}.
A fifth challenge is the expensive data generation process. A series of recent studies remarks that the acquisition of images and especially the labeling of images is costly due to the required expert knowledge and intensive manual work \cite{Ren2017, Tao2018}. Consequently, we derive the fourth design requirement:
\begin{requirement}%[Learn from small amounts of data]
	Industrial surface defect detection systems should be able to learn from small amounts of training data.
\end{requirement}

\section{Rigor Cycle: Drawing on Deep Learning Theory}
\label{sec:rigor}

To address the design requirements derived in the previous section, we identify design principles by drawing on relevant literature as well as insights from domain experts in the field of visual inspection. Design principles are generic capabilities of an artifact through which the design requirements are addressed.

% DP1
Since defects are often very small in relation to the dimensions of the examined part, they can only be captured appropriately by capturing multiple segments of the part rather than photographing the entire part at once. Consequently, we derive the first design principle, which addresses DR1:
\begin{principle}%[Segment-wise Examination]
	Provide the system with segment-wise examination capabilities.
\end{principle}

% DP2
Shang et al. \cite{Shang2018} remark that deep neural nets should be trained on balanced datasets to make more reliable predictions. Oversampling and undersampling are common techniques to adjust the class distribution of a dataset. Oversampling techniques oversample the minority class to create a balanced dataset, and undersampling strategies undersample the majority class to create a balanced dataset. Consequently, we derive the second design principle, which addresses DR2:
\begin{principle}%[Data Balancing]
	Provide the system with data balancing functions.
\end{principle}

% DP3
Several authors address the problem of surface defect detection as a binary classification problem \cite{Soukup2014, Park2016, Tabernik2019}. They argue that an accurate per-image classification is often more important than an accurate localization of the defect. Others address the problem as a multi-class classification problem, where the model has to specify the defect type \cite{Masci2012, Weimer2016, FaghihRoohi2016, Wang2016, Tao2018, Jung2018}. Some authors argue that the precise localization of defects is crucial and address the problem as a segmentation problem \cite{Ren2017, Ferguson2018, Racki2018}. Segmentation models output a visual localization of the defect in the form of a segmentation map, which provide higher information value compared to binary or even multi-class classification. However, there seems to be no consensus on how to address the problem of surface defect detection. Instead, the problem of surface defect detection is addressed according to the goals and priorities of the specific use case. Consequently, we derive the third design principle, which addresses DR3 and DR4:
\begin{principle}%[Information Value]
	Provide the system with mechanisms to address the appropriate information needs of the users based on the objectives and priorities of the specific use case.
\end{principle}

% DP4
Previous research shows that the use of pre-trained weights from large image datasets yields performance improvements over deep neural networks trained from scratch \cite{Ren2017, Shang2018, Jung2018, Ferguson2018}. Consequently, we derive the fourth design principle, which addresses DR4:
\begin{principle}%[Knowledge Transfer]
	Provide the system with knowledge transfer functions to utilize shared features from other models.
\end{principle}

% DP5
Recent studies indicate that regularization methods prevent deep neural networks from overfitting and thus improve model performance. Popular regularization methods include dropout \cite{Wang2016, Ferguson2018, Di2019} and data augmentation \cite{Soukup2014, Weimer2016, Wang2016, Tao2018, Di2019}. Dropout is a technique that randomly drops units and their connections from the network during training \cite{Srivastava2014}. This prevents the units from co-adapting too much. Data augmentation is a technique that increases the diversity of the training set by applying random but realistic transformations such as flipping and rotation. Consequently, we derive the fifth design principle, which addresses DR4:
\begin{principle}%[Regularization]
	Provide the system with regularization mechanisms to to prevent the model from overfitting.
\end{principle}

\section{Design Cycle 1: Strategies for Enabling Deep Learning-based SDDS}
\label{sec:design_cycle1}

\begin{figure*}[h]
	\centering
	\includegraphics[width=0.6\textwidth]{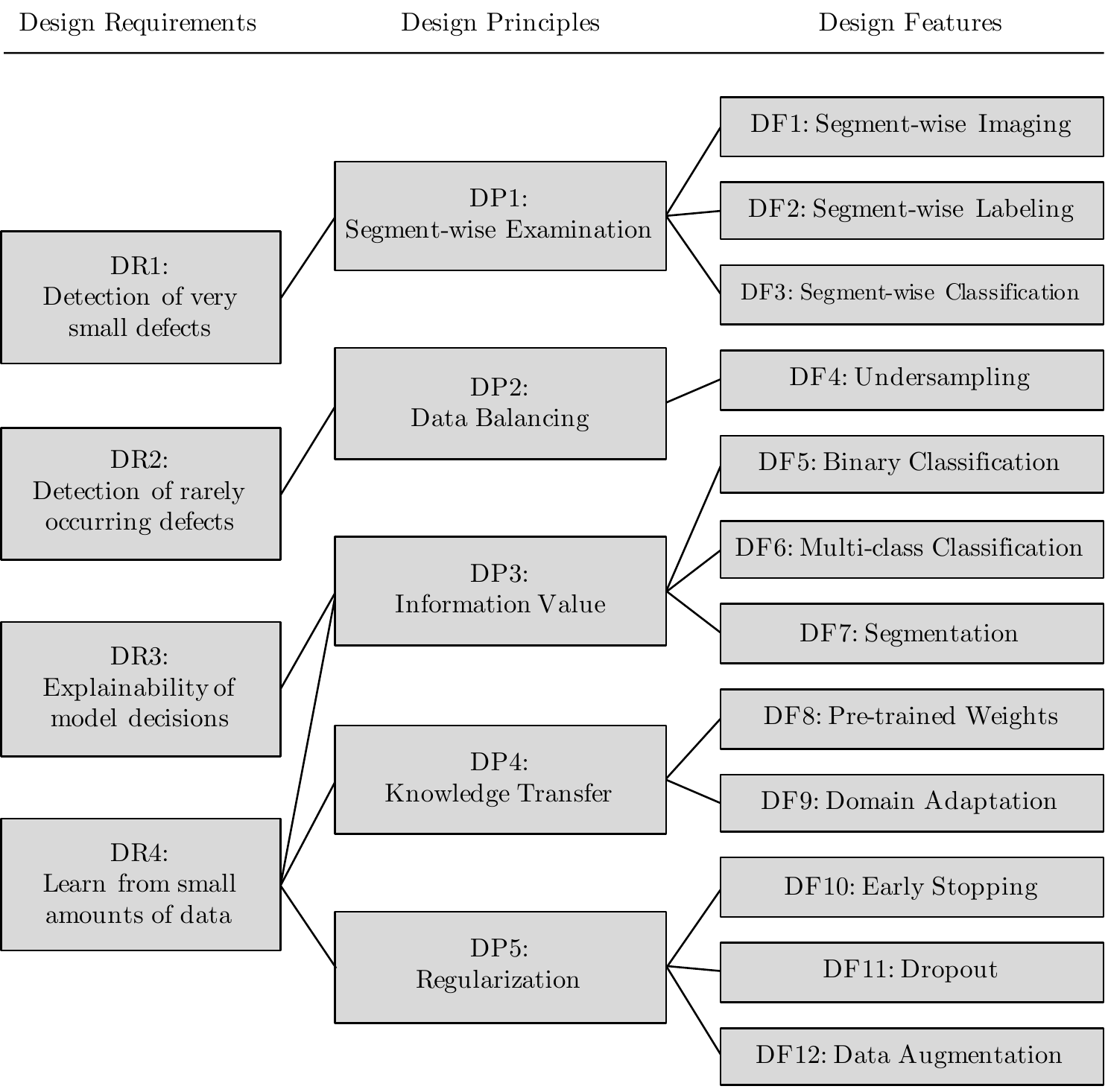}
	\caption{Framework of interrelated design requirements, design principles, and design features}
	\label{fig:mapping}
\end{figure*}

Building on the design principles presented in the previous section, this section derives design features capturing concrete instantiations in the specific context of industrial surface defect detection use cases. \Cref{fig:mapping} depicts an overview of the design features, design principles, and design requirements.

% Figure: Mapping

% DP1
Three design features implement the first design principle. First, we photograph a product in segments (DF1), second, we label each segment/image (DF2) and, third, we derive a product classification based on several individual segment classifications (DF3).
% DP2
The second design principle is addressed by undersampling the majority class to adjust the class distribution so that each class is equally represented (DF4).
% DP3
The third design principle requires defining the appropriate information value and is addressed by three different design features. The model can either output a binary classification (DF5), a multi-class classification (DF6), or a segmentation map (DF7).
% DP4
The fourth design principle postulates knowledge transfer and is implemented by using pre-trained weights (DF8) or a domain adaptation method (DF9). The pre-trained weights come from a dataset like ImageNet or some other large image dataset.
% DP5
The fifth design principle is addressed by stopping the training of a model early when a monitored metric has stopped improving (DF10), randomly dropping units and their connections during training (DF11), and applying random transformations to the training data (DF12).

\section{Design Cycle 2: Surface Defect Detection Prototype Instantiation}
\label{sec:design_cycle2}

To evaluate the applicability and usefulness of the framework on an industrial use case, we conduct eight technical experiments in which we instantiate different deep learning strategies.

\subsection{Industrial Use Case}
\label{sec:use_case}

We evaluate the proposed SDDS in a company, which manufactures so-called engineered molded parts, which are custom-designed components. Currently, these molded parts are 100\% manually inspected before delivery to the customer. Workers inspect each part with a magnifying lens to ensure that it contains no defects. These parts have a much larger diameter than, for example, conventional O-rings and therefore do not have a rigid shape. This makes it difficult to automate the inspection of these engineered molded parts.

The images are taken in a controlled research and development environment. The camera only captures a small segment of the part at a time, and a motor continuously rotates it in front of the camera until every segment of the part has been captured (DF1). This generates 135 slightly overlapping images per part, where each image is a grayscale image with a resolution of about 25 mega pixels.

\begin{figure}[h]
	\centering
	\begin{subfigure}{0.15\textwidth}
		\centering
		\includegraphics[width=\textwidth]{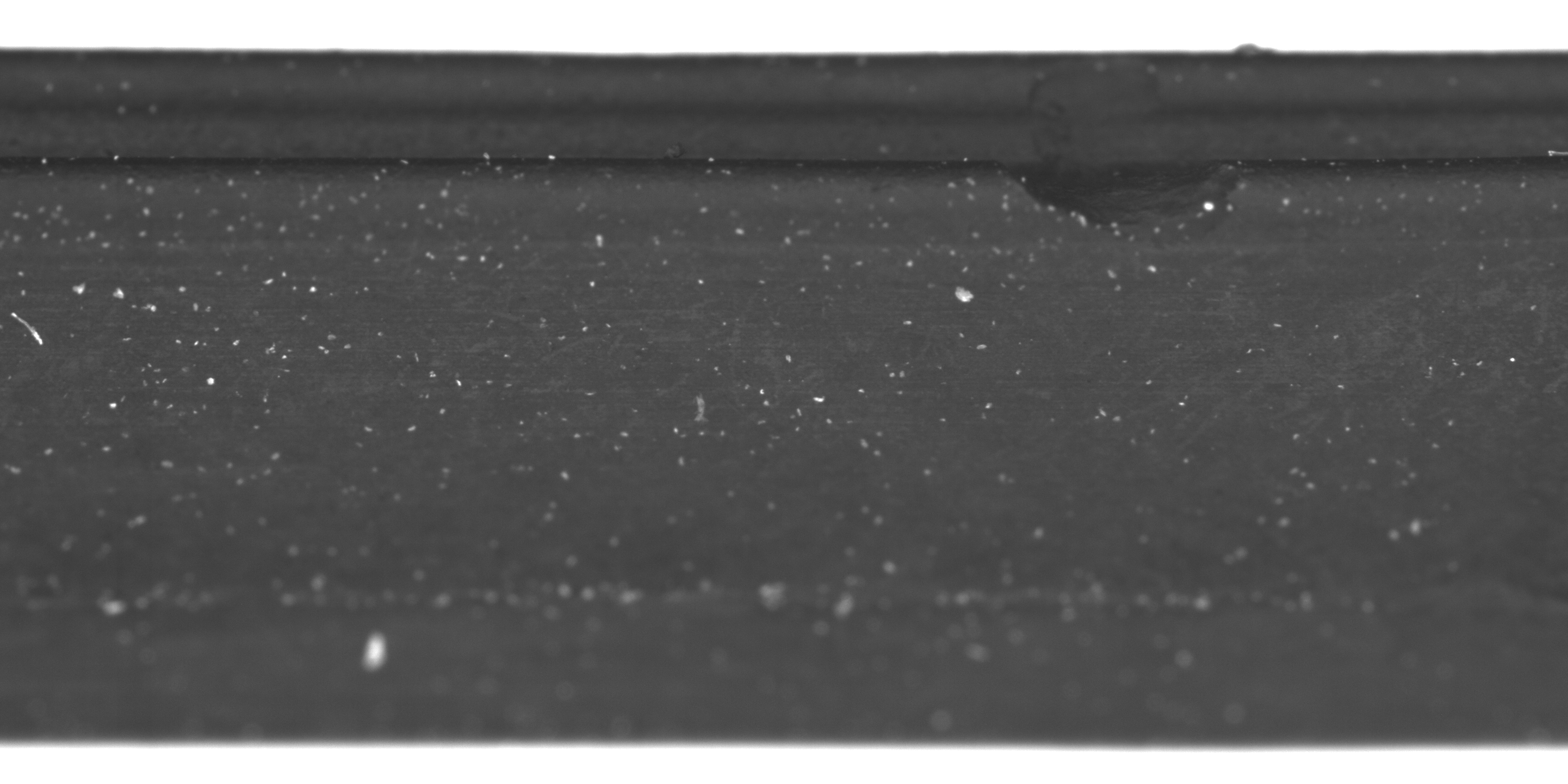} 
		\caption{Nonfill}
		\label{fig:subim1}
	\end{subfigure}
	\hfill
	\begin{subfigure}{0.15\textwidth}
		\centering
		\includegraphics[width=\textwidth]{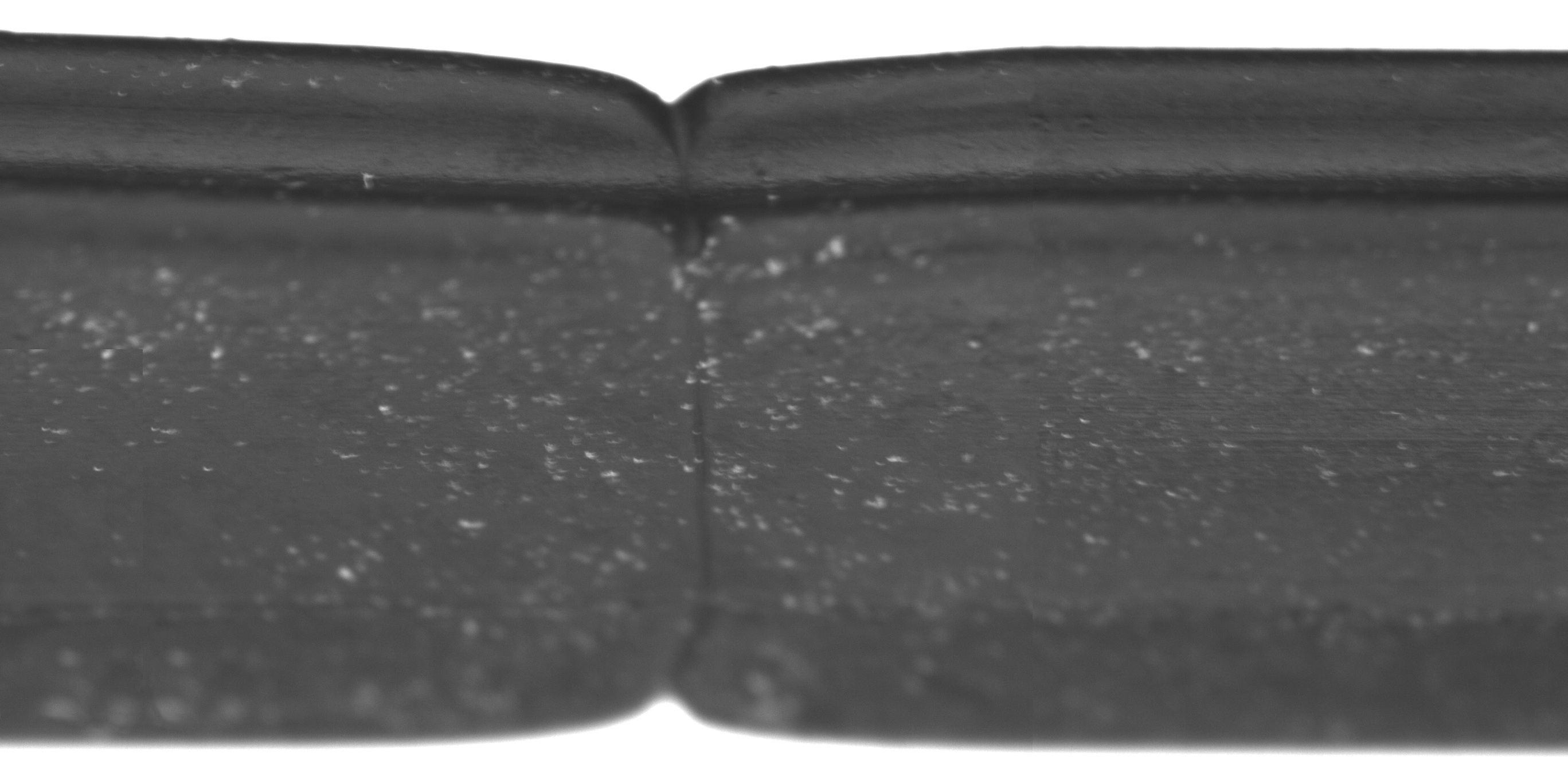}
		\caption{Joining Marks}
		\label{fig:subim3}
	\end{subfigure}
	\hfill
	\begin{subfigure}{0.15\textwidth}
		\centering
		\includegraphics[width=\textwidth]{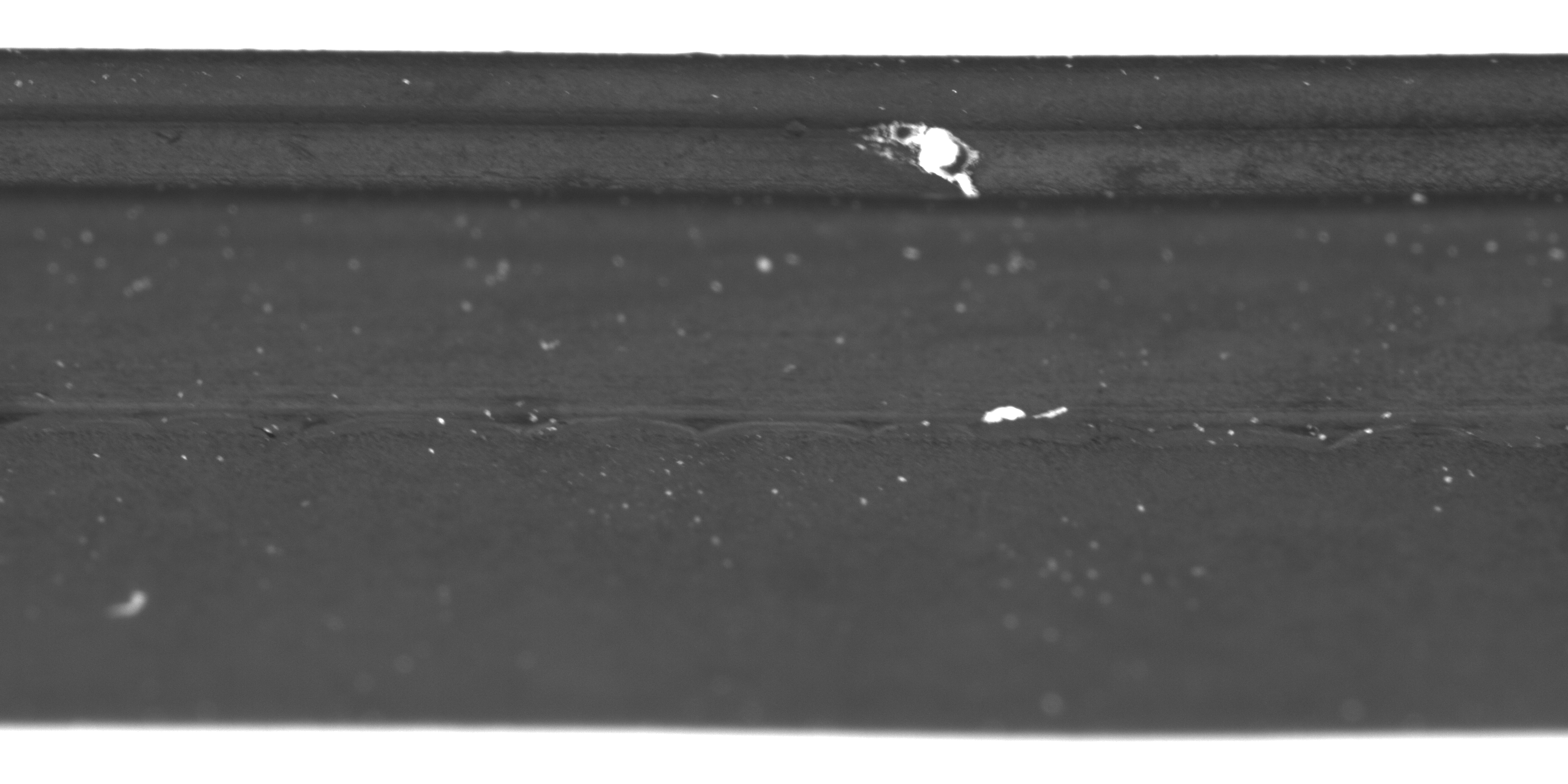}
		\caption{Dirt/Color}
		\label{fig:subim3}
	\end{subfigure}
	\caption{Sample images of different defect types}
	\label{fig:defect_types}
\end{figure}

In this way, we capture 324 defective parts containing five different types of defects, resulting in an initial dataset of 43,740 raw images. 
%\Cref{tab:sample_count} gives an overview of the number of partss and captured images per defect type, and 
\Cref{fig:defect_types} shows three exemplary defect types. 
%To reduce the images' file size, we remove excess white space in the images by cropping. 
Most parts are only defective at one location, so most of the segments are considered non-defective. This results in a very unbalanced dataset, where only about 1.5\% of images contain a defect. Therefore, we apply an undersampling strategy to balance out the class distribution (DF4). This ensures that the model does not overfit the majority class and sees defective and non-defective images at the same frequency. The final dataset consists of 1.280 images.

% Table: Sample count
%\begin{table}[h]
%	\caption{Number of samples, images, and defect images of each defect type}
%	\label{tab:sample_count}
%	\centering\small
%	\begin{tabular}{llll}
%	\hline
%	Defect Type       & Samples & Images & Defects \\ 
%	\hline
%	Nonfill       & 100     & 13.500 & 210                \\
%	Crack         & 50      & 6.750  & 103                \\
%	Joining Marks & 87      & 11.745 & 162                \\
%	Dirt/Color    & 37      & 4.995  & 67                 \\
%	Brush Damage  & 50      & 6.750  & 98                 \\ 
%	\hline
%	Total             & 324     & 43.740 & 640            %    \\ 
%	\hline	
%	\end{tabular}
%\end{table}

% Figure: Defect type images

\subsection{Technical Experiments}
\label{sec:technical_experiment}

The main focus of the experiments is to investigate deep learning strategies based on DP3 and DP4. On the one hand, deep learning models can provide different amounts of \emph{information value} (DP3) and, on the other hand, they can use different amounts of \emph{knowledge transfer} (DP4). The information value refers to the output of a deep learning model and impacts the explainability and comprehensibility of the model decision. A model can predict whether an image contains a defect or not (binary classification), which defect type it contains (multi-class classification) or where the defect is located in the image (segmentation). However, we hypothesize that an increase in information value is associated with higher learning difficulty, which leads to decreased model performance. Knowledge transfer relates to the amount of abstract knowledge being transferred from other tasks or domains. A model can be trained solely on the data collected for the task, or it can utilize knowledge from a model that has been trained on a generic dataset like ImageNet \cite{Russakovsky2015} (generic transfer). Another option is to transfer knowledge from a model that has been trained on an industrial dataset that is supposedly more similar to the target dataset than a generic dataset (industrial transfer). We hypothesize that an increase in knowledge transfer leads to better model performance and shorter training times.

\begin{figure*}[h]
	\centering
	\includegraphics[width=0.7\textwidth]{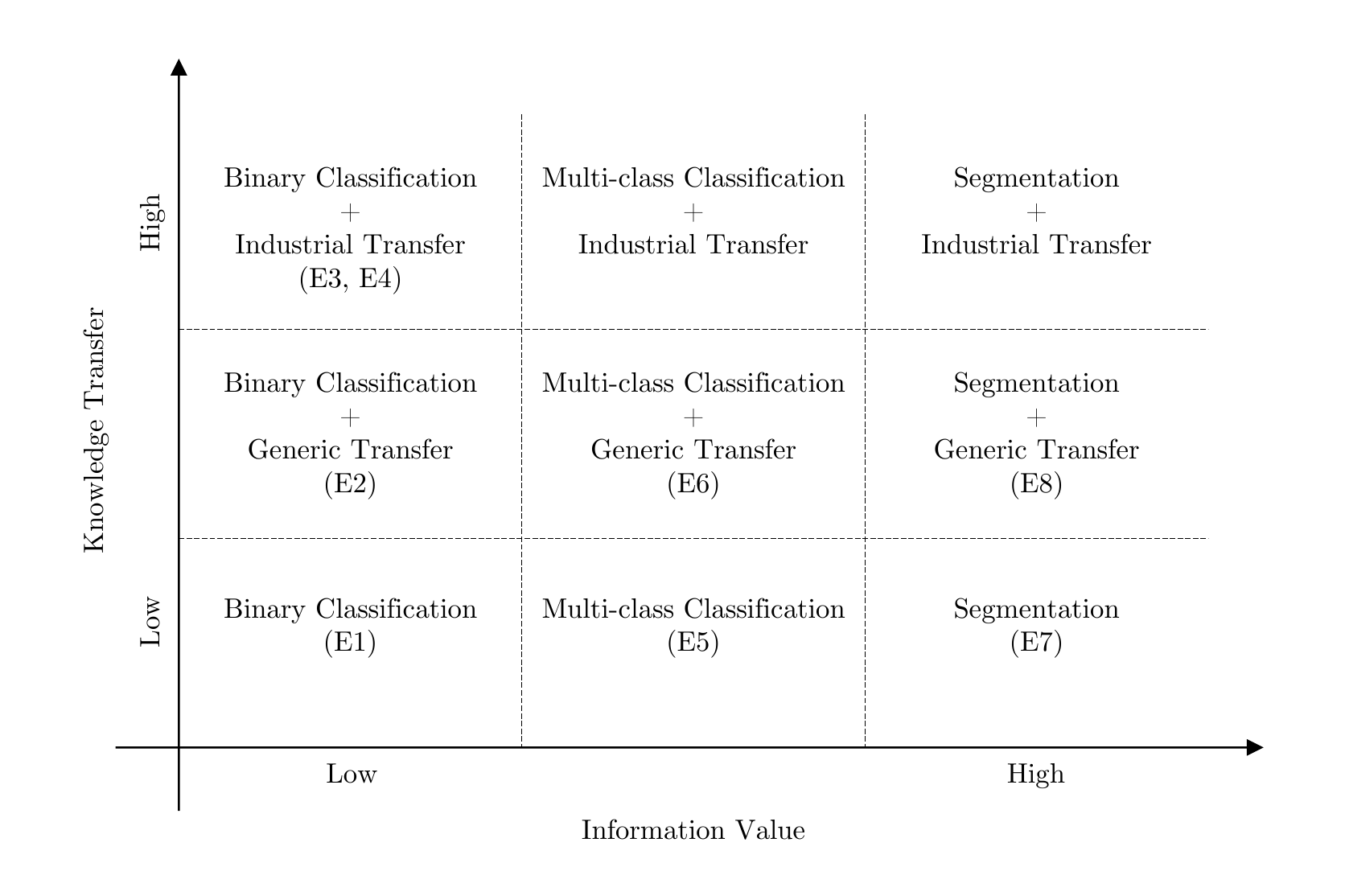}	
	\caption{Overview of experimental scenarios addressing information value and knowledge transfer levels}
	\label{fig:experimental_scenarios}
\end{figure*}

Combining these two dimensions leads to nine different experimental scenarios (\Cref{fig:experimental_scenarios}). We conduct eight experiments in which we cover seven of the scenarios. The two remaining scenarios are not covered due to the lack of an appropriate industrial dataset. The binary classification with industrial transfer scenario is covered by two experiments using two different transfer approaches (DF8 and DF9). However, all experiments also cover the remaining design principles and features from the framework (\Cref{tab:design_features}).

% Figure: Experimental scenarios

% Table: Overview of experiments
\begin{table*}[h]
	\caption{Overview of the experiments and the implemented design features}
	\label{tab:design_features}
	\centering\small
	\resizebox{\textwidth}{!}{
		\begin{tabular}{lcccc ccccc ccc}
		\hline
		Experiment & DF1 & DF2 & DF3 & DF4 & DF5 & DF6 & DF7 & DF8 & DF9 & DF10 & DF11 & DF12 \\ 
		\hline
		E1         & x   & x   & x   & x   & x   & -   & -   & -   & -   & x    & x    & -    \\
		E2         & x   & x   & x   & x   & x   & -   & -   & x   & -   & x    & x    & x    \\
		E3         & x   & x   & x   & x   & x   & -   & -   & x   & -   & x    & x    & x    \\
		E4         & x   & x   & x   & x   & x   & -   & -   & -   & x   & -    & x    & x    \\
		E5         & x   & x   & x   & x   & -   & x   & -   & -   & -   & x    & x    & -    \\
		E6         & x   & x   & x   & x   & -   & x   & -   & x   & -   & x    & x    & x    \\
		E7         & x   & x   & x   & x   & -   & -   & x   & -   & -   & x    & -    & -    \\
		E8         & x   & x   & x   & x   & -   & -   & x   & x   & -   & x    & -    & -    \\
		\hline
		\end{tabular}}
\end{table*}

% Binary (E1, E2, E3, E4)
In the first three experiments (E1, E2, E3), we train binary classification models using different knowledge transfer strategies. The models build on a modified ResNet50 \cite{He2016} architecture and consist of five blocks of convolutional layers, a global average pooling layer, a dropout layer (DF11), and a fully-connected layer for binary classification (DF5). In experiment E1, we do not transfer any knowledge and initialized the weights randomly. In experiment E2, we apply a generic knowledge transfer by using pre-trained weights from ImageNet \cite{Russakovsky2015} (DF8). In experiment E3, we apply an industrial knowledge transfer by using pre-trained weights from a model trained on a Kaggle dataset \cite{Severstal2019} (DF8). This dataset consists of 12,586 images of steel defects and contains four types of defects. We chose this dataset because of its relatively large size, visual similarity, and public availability. In experiments E2 and E3, we apply data augmentation by randomly flipping, zooming, and shifting the images (DF12).

In experiment E4, we use a modified CycleGAN \cite{Zhu2017} to train a network that transforms images of rubber parts into steel images (DF9). We then use this network and a binary classifier trained on the steel dataset to detect rubber part defects. 
%The CycleGAN architecture consists of two generator networks and two discriminator networks. The generators learns to transform rubber images into steel images and vice versa, while the discriminators learns to differentiate between real rubber/steel images and generated rubber/steel images. The generators are modified U-Nets \cite{Ronneberger2015}, consisting of a contraction path and an expansion path. The contraction path, also referred to as the encoder, reduces the input's spatial information while increasing the feature information. The expansion path or decoder consists of a series of transposed convolutions and concatenation with the contraction path's corresponding high-resolution features. It also includes a dropout layer (DF11) for regularization. The discriminators are PatchGANs \cite{Isola2017} and only penalize structure at the scale of local image patches. They try to classify each $N \times N$ patch in an image as real or fake. The binary classifier is a modified ResNet50 like in the first three experiments.

% Multi (E5, E6)
In experiments E5 and E6, we train multi-class classification models using different amounts of knowledge transfer. We use the same architecture as in the first three experiments, except that we modify the output layer for a multi-class classification with six classes (DF6). In experiment E5, we do not transfer any knowledge and initialize the weights randomly. In experiment E5, we apply a generic knowledge transfer by using pre-trained weights from ImageNet (DF8). In experiment E6, we apply the same data augmentation transformations as in experiments E2 and E3 (DF12).

% Segmentation (E7, E8)
In experiments E7 and E8, we train segmentation networks using different amounts of knowledge transfer. The output of these networks is a pixel-wise mask of the input image, indicating which pixels belong to which class (DF7). The networks are modified U-Nets \cite{Ronneberger2015}.

\subsection{Evaluation}
\label{sec:evaluation}

To better compare the different deep learning strategies, the problem of surface defect detection is translated into a binary classification problem. We evaluate the multi-class classification models in a one-vs-all fashion and transform the segmentation models' output segmentation masks into binary classifications. We use a simple thresholding method, where a segmentation mask is considered a positive classification if the sum of the pixel values of the segmentation mask is above a certain threshold. The threshold value is set to achieve the best possible classification accuracy in each experiment, respectively.

For all experiments, we report accuracy, precision and recall as well as $F_{1}$ score as the primary evaluation metric (\Cref{tab:metrics}). Note that the $F_{1}$ score is chosen because it more accurately captures a classifier's performance on unbalanced datasets. This is relevant for the multi-class classification models since the different defect types are represented unevenly. %Additionally, we plot the Receiver Operating Characteristic (ROC) curves (\Cref{fig:roc}) and report the corresponding area under the curve (AUC). The ROC curve plots the true positive rate against the false positive rate at various threshold settings. Thus, it captures the trade-off between recall and precision. The AUC measures the area under the ROC curve and enables the comparison of different ROC curves through a single value.

% Table: Binary metrics
\begin{table}[h]
	\caption{Performance metrics of binary, multi-class classification and segmentation experiments}
	\label{tab:metrics}
	\centering\small
	\begin{tabular}{l ccccc}
		\hline
		Experiment & Accuracy & Precision & Recall & $F_{1}$ score    \\
		\hline
		E1         & 0.664    & 0.666     & 0.664  & 0.665         \\
		E2         & 0.974    & 0.974     & 0.974  & 0.974         \\
		E3         & \textbf{0.977}    & \textbf{0.977}     & \textbf{0.977}  & \textbf{0.977}          \\
		E4         & 0.500      & 0.250      & 0.500    & 0.333          \\
		\hline
		E5         & 0.549    & 0.338     & 0.276  & 0.28         
		\\
		E6         & 0.930     & 0.903     & 0.894  & 0.898         
		\\
		\hline
		E7         & 0.698    & 0.702     & 0.698  & 0.697         \\
		E8         & 0.930     & 0.930      & 0.930   & 0.930           \\
		\hline
	\end{tabular}
\end{table}

% Figure: Classification ROC curves
%\begin{figure*}[h]
%	\centering
%	\begin{subfigure}{0.32\textwidth}
%		\centering
%		\includegraphics[width=\textwidth]{figures/binary_roc.png}
%		\caption{Binary classification}
%		\label{subfig:binary_roc}
%	\end{subfigure}
%	\hfill
%	\begin{subfigure}{0.32\textwidth}
%		\centering
%		\includegraphics[width=\textwidth]{figures/multi_roc.png}
%		\caption{Multi-class classification}
%		\label{subfig:multi_roc}
%	\end{subfigure}
%	\hfill
%	\begin{subfigure}{0.32\textwidth}
%		\centering
%		\includegraphics[width=\textwidth]{figures/seg_roc.png}
%		\caption{Segmentation}
%		\label{subfig:segment_roc}
%	\end{subfigure}
%	\caption{ROC curves of binary, multi-class classification and segmentation experiments}
%	\label{fig:roc}
%\end{figure*}

Considering the binary experiments, E3 (industrial transfer) outperforms the other models with a score of 0.977 in all metrics.
%\Cref{subfig:binary_roc} shows the ROC curves of the models. 
%However, experiment E2 achieves the highest AUC score with 0.998. Additionally, experiments E2 and E3 converge faster than E1.
%Experiment E4 transforms rubber images into steel images using a CycleGAN architecture. However, it only achieves an $F_{1}$ score of 0.333 and, therefore, can not keep up with the other experiments.
%Considering multi-class experiments, E6 (generic transfer) outperforms E5 (no transfer) in performance as well as convergence behaviour.
%Similarly, considering segmentation experiments isolated, E8 performs better than E7 and also converges faster than the model in experiment E7. 
%\Cref{fig:seg_output} compares the segmentation results of E7 and E8 on exemplary defects.

% Figure: Seg outputs
\begin{figure}[h]
	\centering
	\includegraphics[width=0.48\textwidth]{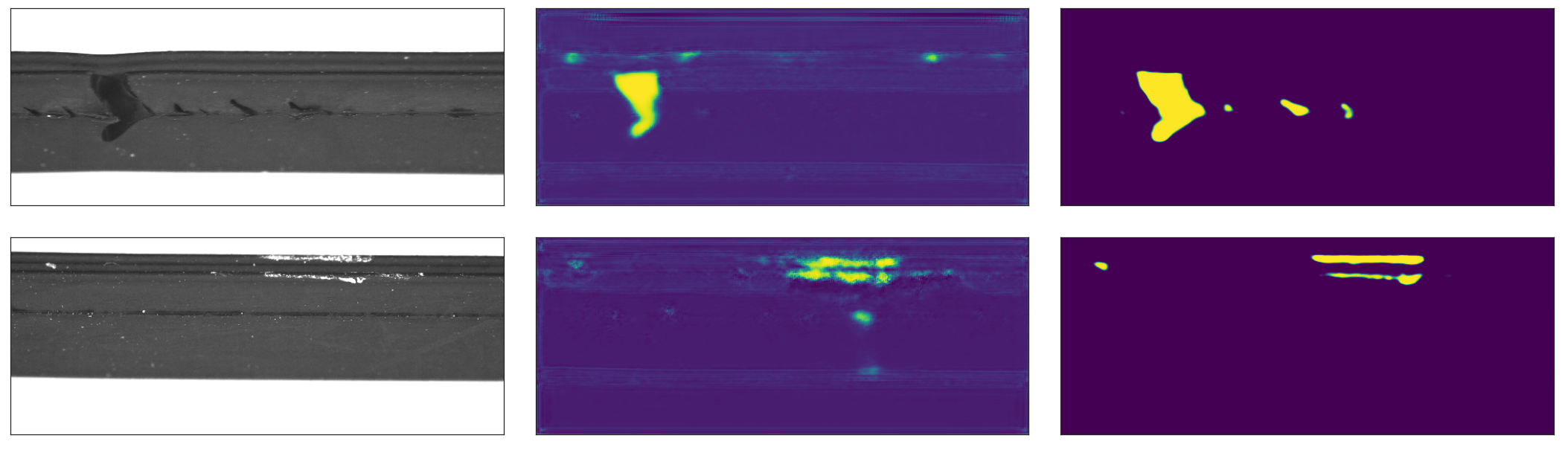}
	\caption{Example inputs (left) and the corresponding predicted segmentation masks in segmentation experiment E7 (middle) and E8 (right)}
	\label{fig:seg_output}
\end{figure}

Overall, binary classification performs slightly better than multi-class classification as well as segmentation. However, when focusing on the $F_{1}$ score, segmentation outperforms multi-class classification.

\section{Discussion}
\label{sec:discussion}

By instantiating the proposed design features and their respective design principles through the conducted experiments, applicability and usefulness of the deep learning strategies framework on an industrial surface defect detection tasks is demonstrated. In the following, we separately discuss design principles and their impact. 

%\subsection{Segment-wise Examination and Data Balancing}
To the best of our knowledge, this article is the first one applying deep learning methods in the context of surface defect detection of engineered molded parts. The characteristics of these rubber parts present unique challenges to the application of deep learning methods. The very small size of the defects compared to the parts' large surface constitutes a major challenge when capturing the defects with cameras. We address this issue by \emph{segment-wise examination (DP1)}. In particular we capture multiple segments (DF1) for each part and label (DF2) and classify (DF3) the segments separately. This enables us to achieve the required image quality for the detection of surface defects. We also apply an undersampling strategy (DF4) to \emph{balance the dataset (DP2)}.

%\subsection{Information Value}
Furthermore, before conducting the experiments, we hypothesised that an increase in \emph{information value (DP3)} is associated with higher learning difficulty, which leads to decreased model performance. In experiments E2, E6, and E8, we investigate different information values by implementing either a binary classification model (DF5), a multi-class classification model (DF6), or a segmentation model (DF7) with generic knowledge transfers. By comparing the results of these experiments, we see that the model performance decreases when comparing the binary classification (E2) to the multi-class classification model (E6). However, the segmentation model (E8) achieves the same accuracy score as the multi-class classification model and reaches even higher precision, recall, and $F_{1}$ scores. A possible explanation for these results lies in the nature of the task itself. The problem of image segmentation is essentially a problem of image classification on the pixel level. In image segmentation, every pixel of an image is associated with a label and constitutes a training sample to the algorithm. Thus, with the same number of training images, the segmentation model can access more training samples and receives more expert knowledge than the binary or multi-class classification models. The larger amount of training samples or label information might have outweighed higher learning difficulty. For practitioners, this would mean that there is no direct trade-off between model performance and information value and that it could actually be beneficial to increase the information value provided by a model without performance loss.

%\subsection{Knowledge Transfer}

\begin{figure}[h]
	\centering
	\includegraphics[width=0.48\textwidth]{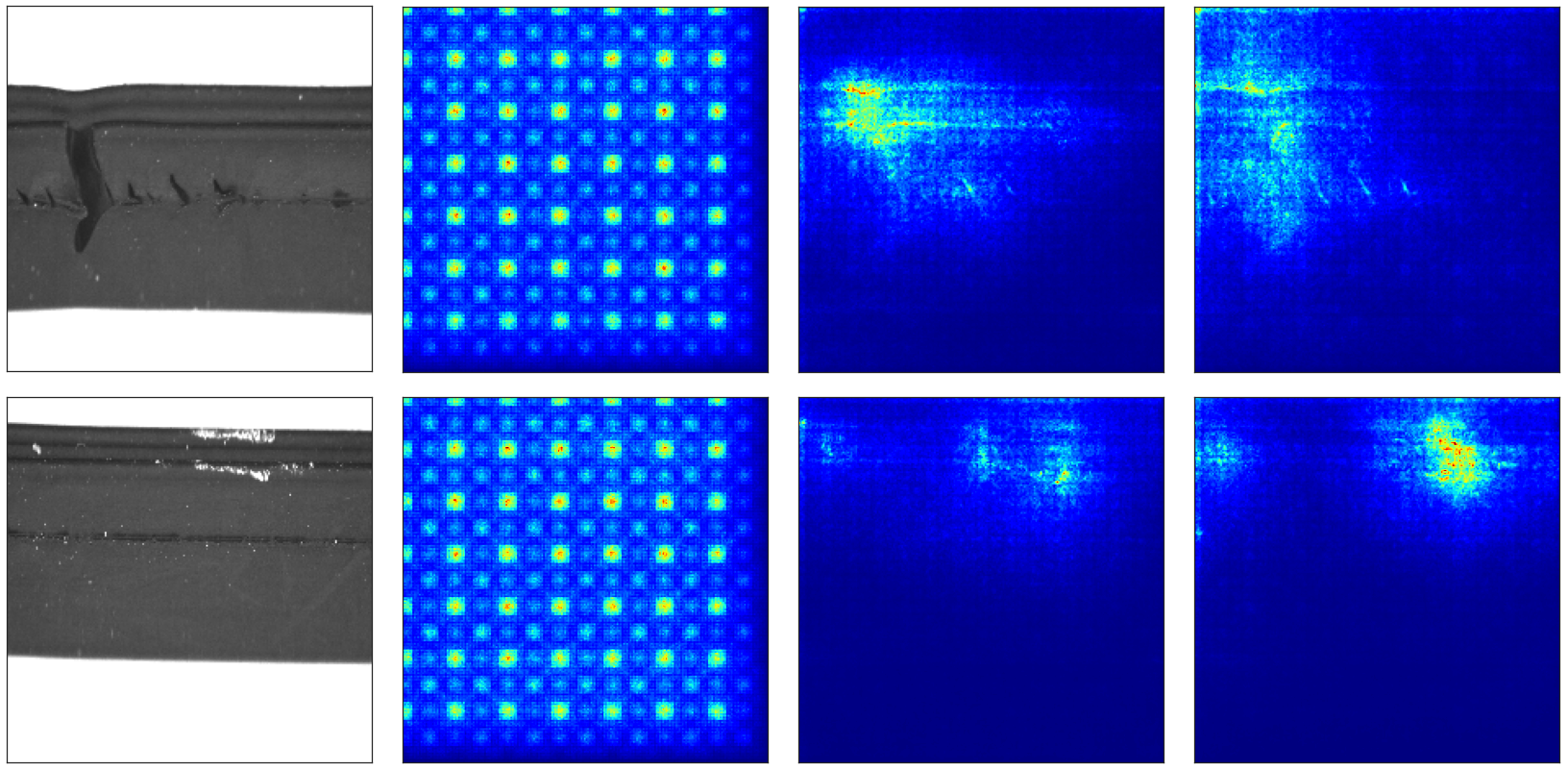}
	\caption{Activation maps of binary classification models E1-E3 (from left to right)}
	\label{fig:binary_saliency}
\end{figure}

We also hypothesized that an increase in \emph{knowledge transfer (DP4)} leads to better model performance as well as faster training times. From the results presented, it is clear that the use of pre-trained weights (DF8) impacts model performance and training time. In the binary classification, the multi-class classification, and the segmentation experiments, the models using pre-trained weights (E2, E3, E6, E8) outperform the models without knowledge transfer (E1, E5, E7). Additionally, the models with pre-trained weights converge faster than the models trained from scratch. \Cref{fig:binary_saliency} shows saliency maps \cite{Simonyan2013} of the binary classification models for five exemplary samples. 
%Saliency maps can be interpreted as heatmaps, where hotness corresponds to regions of an image that greatly impact the models' final decision. 
We can see that the model without knowledge transfer (second column) did not learn to detect defects but instead pays attention to some other pattern in the data. The models with knowledge transfer (third and fourth column) learned to recognize defects correctly. The pre-trained weights from a knowledge transfer leverage additional amounts of training data and produce more sensitive gradients than randomly initialized weights. This helps the model to converge towards the global minimum faster.
However, the binary classification experiments results suggest that an industrial knowledge transfer (E3) offers only a marginal improvement over a generic knowledge transfer (E2). This might be due to the steel dataset not being similar enough to the rubber part dataset. Even though both datasets contain industrial surface defects, the steel and rubber part images' visual appearance is still quite different. Therefore, the steel dataset might not contain significantly more domain-specific knowledge than the generic ImageNet dataset. From this standpoint, the results can be considered a positive indicator that an actual industrial knowledge transfer, for example, from one molded part type to another one, can produce more significant performance improvements. This assumption should be addressed in future research. For practitioners, the key finding is that already a generic knowledge transfer leads to significant performance improvements. 

% Figure: Binary saliency

%Moreover, the results of experiment E4 indicate that the domain adaptation approach (DF9) using a CycleGAN architecture does not work as intended. At this stage of understanding, we believe that the problem lies in the generator not knowing how to transform a rubber part defect into a steel defect. In our architecture, the generator learns to transform the general domain of rubber images into steel images. This could mean that a positive rubber part image is transformed into a negative steel image. Therefore, it remains unclear whether this type of approach can yield usable results, and we have to conclude that this feature does not successfully address design requirement DR4. Nevertheless, the advantages of such an approach are quite obvious, as this approach would alleviate the need for labeled data in the target domain, and therefore, future research on this topic seems significant.

%\subsection{Regularization}

In all experiments, we see that training and validation loss converge together and that the validation loss does not increase again in any of the experiments. Therefore, we conclude that the applied \emph{regularization (DP5)} techniques, such as early stopping (DF10), dropout (DF11), and data augmentation (DF12) features successfully prevent the models from overfitting and address design requirement DR4.

\section{Conclusion}
\label{sec:conclusion}

% Summary
This article utilizes a design science research approach to investigate suitable strategies that enable the successful application of deep learning methods in industrial surface defect detection systems. More specifically, we conceptualized a framework of interrelated design requirements, design principles, and design features that captures suitable deep learning strategies for industrial SDDS. In a series of experiments, we utilized the framework to build different deep learning models in an industrial case study. We achieved a 97.7\% accuracy in the binary classification of molded part defects using only a very small dataset. The evaluation results showed that transferring knowledge from a generic dataset significantly improves the performance of models for industrial applications. Furthermore, the results indicated that deep learning methods can be successfully applied in surface defect detection systems and that our framework provides a set of practical guidelines for developing visual inspection solutions.

% Limitations
The results, however, should be assessed in light of its limitations. 
A first limitation relates to the experimental evaluation of deep learning strategies in a single use case. While a quantitative and broader investigation of deep learning strategies is desirable and encouraged, we want to emphasize that while writing this article, there were no sufficiently large and labeled datasets publicly available for most industrial applications.
A second limitation refers to the variety and number of conducted experiments. Our experiments are focused primarily on two aspects of the framework. Conducting further experiments would have enabled us to draw more substantiated conclusions about the remaining aspects of the framework.
A third limitation relates to the execution of the experiments. Our hyperparameters are based on pre-tests and state-of-the-art recommendations. Conducting a systematic hyperparameter optimization might have resulted in slightly better model performances; however, our main goal is to evaluate different deep learning strategies and not to achieve the best model performance possible.

% Future research
Beyond the aforementioned opportunities, there are many other possibilities to extend the work of this article. We encourage scholars to further investigate the impact of information value on model performance and the amount of required training data. Another interesting opportunity lies in the further investigation of industrial knowledge transfers with more suitable datasets.

% if added before the last page, this command can help balancing columns
%\addtolength{\textheight}{-.2cm} 

%Bibliography 
\bibliographystyle{ieeetr}
\bibliography{references}

\end{document}